\documentclass[runningheads]{llncs}
\usepackage{graphicx}
\usepackage{amsmath,amssymb} 
\usepackage{color}
\newcommand{\sminus}{\scalebox{0.75}[1.0]{-}}
\newcommand{\splus}{\scalebox{0.75}[1.0]{+}}
\usepackage{booktabs}       
\usepackage{amsfonts}       
\usepackage{subcaption}
\usepackage{multirow}
\usepackage{dsfont} 
\usepackage{hyperref}
\usepackage[capitalise,nameinlink]{cleveref}
\renewcommand{\ref}[1]{\autoref{#1}}

\def\imagetop#1{\vtop{\null\hbox{#1}}} 
\usepackage[font=footnotesize,labelfont=bf,skip=0.2pt]{caption}
\usepackage{graphbox} 

\usepackage{tikz}
\usetikzlibrary{tikzmark}

\begin{document}
\title{ContextVP: Fully Context-Aware Video Prediction} 

\titlerunning{ContextVP: Fully Context-Aware Video Prediction}
%
\author{Wonmin Byeon\inst{1,2,3,4} \and
Qin Wang\inst{2} \and \\
Rupesh Kumar Srivastava\inst{4} \and 
Petros Koumoutsakos\inst{2}} 
%
\authorrunning{W. Byeon, Q. Wang, R. K. Srivastava, and P. Koumoutsakos}
%

\institute{NVIDIA, Santa Clara, CA, USA \\ 
\email{wbyeon@nvidia.com} \and
ETH Zurich, Zurich, Switzerland \and
The Swiss AI Lab IDSIA, Manno, Switzerland \and
NNAISENSE, Lugano, Switzerland 
}
%
\maketitle              
\begin{abstract}
Video prediction models based on convolutional networks, recurrent networks, and their combinations often result in blurry predictions.
We identify an important contributing factor for imprecise predictions that has not been studied adequately in the literature: blind spots, i.e., lack of access to all relevant past information for accurately predicting the future.
To address this issue, we introduce a fully context-aware architecture that captures the entire available past context for each pixel using Parallel Multi-Dimensional LSTM units and aggregates it using blending units.
Our model outperforms a strong baseline network of 20 recurrent convolutional layers and yields state-of-the-art performance for next step prediction on three challenging real-world video datasets: Human 3.6M, Caltech Pedestrian, and UCF-101.
Moreover, it does so with fewer parameters than several recently proposed models, and does not rely on deep convolutional networks, multi-scale architectures, separation of background and foreground modeling, motion flow learning, or adversarial training.
These results highlight that full awareness of past context is of crucial importance for video prediction.
\end{abstract}
\section{Introduction}
\label{sec:intro}
Unsupervised learning from unlabeled videos has recently emerged as an important direction of research.
In the most common setting, a model is trained to predict future frames conditioned on the past and learns a representation that captures information about the appearance and the motion of objects in a video without external supervision.
This opens up several possibilities: the model can be used as a prior for video generation, it can be utilized for model-based reinforcement learning \cite{suba98},
or the learned representations can be transferred to other video analysis tasks such as action recognition~\cite{srivastava2015unsupervised}.
However, learning such predictive models for natural videos is a rather challenging problem due to the diversity of objects and backgrounds, various resolutions, object occlusion, camera movement, dynamic scene and light changes between frames.
As a result, current video prediction models based on convolutional networks, recurrent networks, and their combinations often result in imprecise (blurry) predictions.
Even very large, powerful models trained on large amounts of data can suffer from fundamental limitations that lead to blurry predictions.
The structure of certain models may be inappropriate for the task, resulting in training difficulties and poor generalization.
Some researchers have proposed to incorporate motion priors and background/foreground separation into model architectures to counter this issue.

Blurry predictions are fundamentally a manifestation of model uncertainty, which increases if the model fails to sufficiently capture relevant past information.
Unfortunately, this source of uncertainty has not received sufficient attention in the literature. 
Most current models are not designed to ensure that they can properly capture all possibly relevant past context.
This paper attempts to address this gap.
\begin{figure}[t]
    \centering
        \includegraphics[trim={1.0cm 0cm 1.0cm 0cm},clip,width=0.8\textwidth]{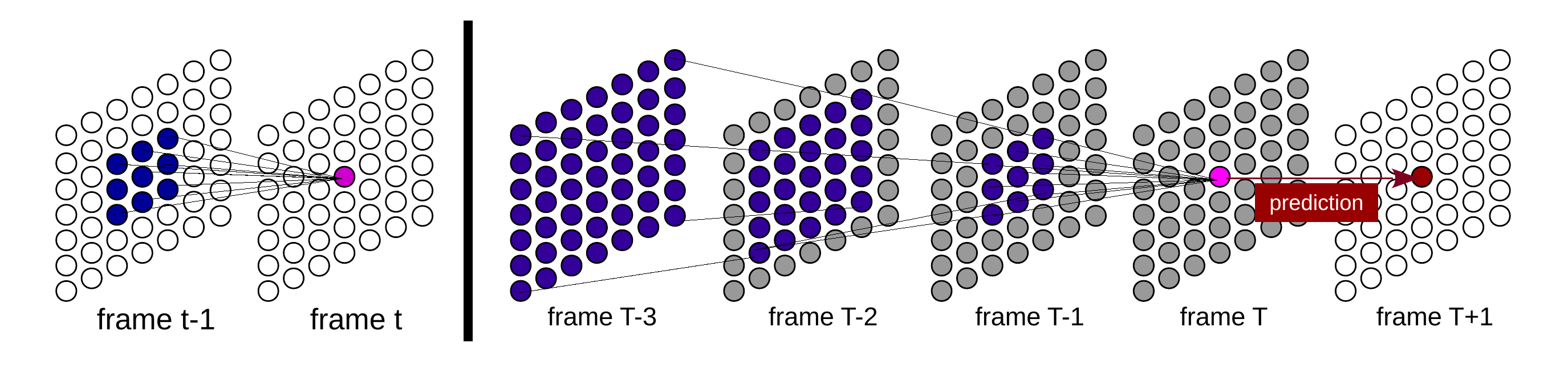}   
\caption{(left) The Convolutional LSTM (ConvLSTM) context dependency between two successive frames. (right) The context dependency flow in ConvLSTM over time for frame $t=T$. Blind areas shown in gray cannot be used to predict the pixel value at time $T+1$. Closer time frames have larger blind areas.}
\label{fig:convlstm}
\end{figure}
\begin{figure}[t]
    \centering
    \begin{subfigure}{\textwidth}
        \includegraphics[trim={1.0cm 0cm 0.5cm 0cm},clip,width=\textwidth]{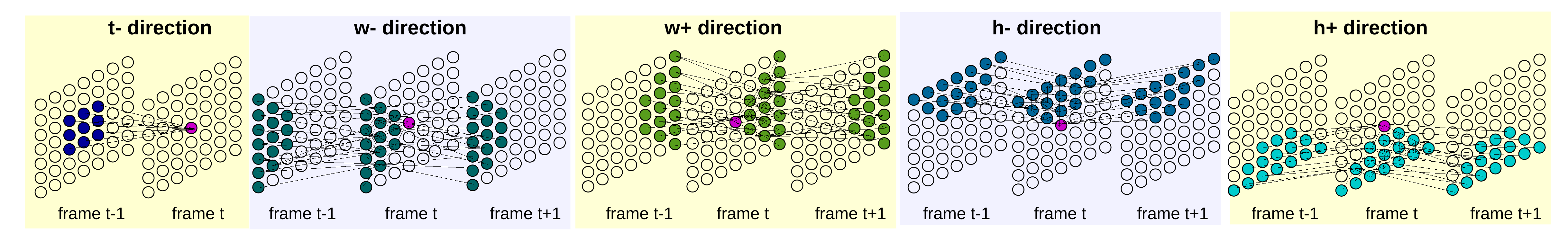}
    \end{subfigure}
    \begin{subfigure}{0.6\textwidth}
        \includegraphics[trim={0cm 0.0cm 0cm 0.8cm},clip,width=\textwidth]{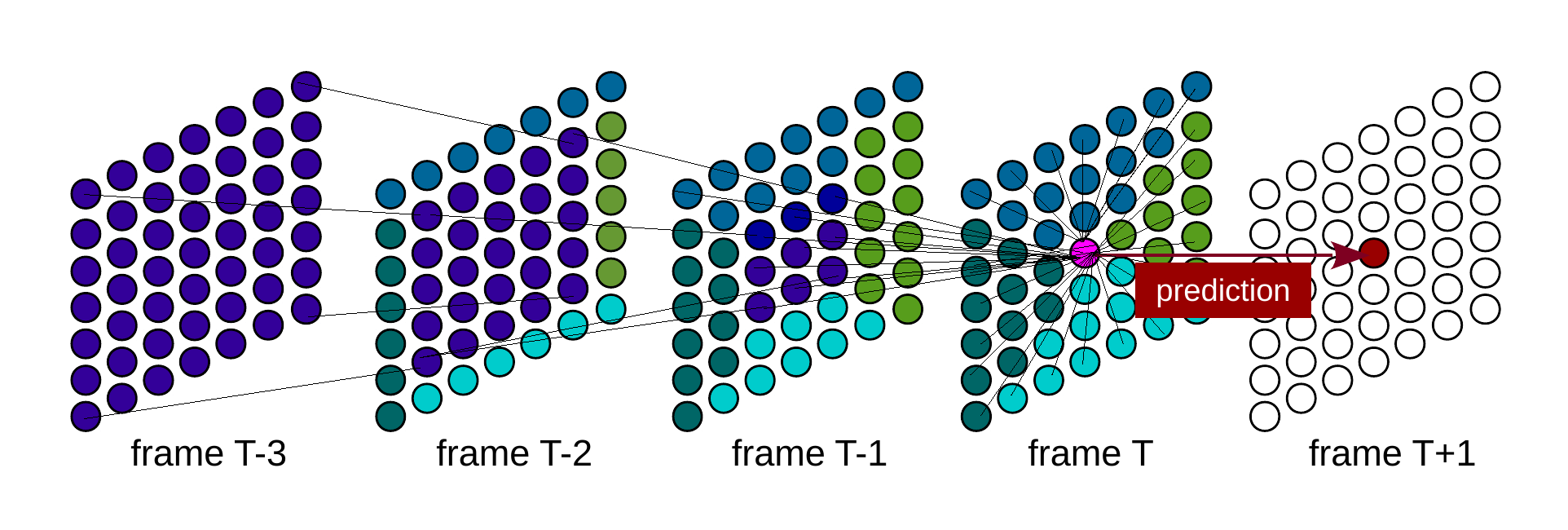}
    \end{subfigure}
\caption{(top) Context dependency between two frames when using Parallel MD-LSTM (PMD) units for five directions: $t-$, $w-$, $w+$, $h-$, and $h+$, where $h$, $w$, and $t$ indicate the current position for height, width, and time dimensions. (bottom) The combined context dependency flow for frame $t=T$ in the proposed architecture. All available context from past frames is covered in a single layer regardless of the input size. 
}
\label{fig:pyramid}
\end{figure}

Our contributions are as follows:
\begin{itemize}
  \item We highlight a \textbf{blind spot problem} in common video prediction models, showing that they do not systematically take the entire spatio-temporal context from past frames into account (see \autoref{fig:convlstm}-right) and have to rely on increasing depth to do so.
  This increases uncertainty about the future which can not be remedied using special loss functions or motion priors.
  \item We contribute a \textbf{simple baseline model that outperforms rather complex models from recent literature}. Due to increased depth, this baseline model has an increased ability to capture relevant context.
  \item We propose a \textbf{new architecture for video prediction} that systematically and efficiently aggregates contextual information for each pixel in all possible directions (left, right, top, bottom, and time directions) at each processing layer (see \autoref{fig:pyramid}) instead of stacking layers to cover the available context.
  We additionally propose \emph{weighted context-blending blocks} and \emph{regularization via directional weight sharing} for the proposed architecture. 
  We obtain performance improvements over our strong baseline as well as state-of-the-art models while using fewer parameters and simple loss functions.
\end{itemize}
We demonstrate improvements in a variety of challenging video prediction scenarios: car driving, human motion, and diverse human actions in YouTube videos.
Quantitative improvements on metrics are accompanied by results of high visual quality showing sharper future predictions with reduced blur or other motion artifacts. 
Since the proposed models do not require separation of content and motion or novel loss functions to reach the state of the art, we find that full context awareness is a crucial ingredient for high quality video prediction.
Code and video results for this paper are available on the project website: \url{https://wonmin-byeon.github.io/publication/2018-eccv}.

\section{Related Work}
\label{sec:relatedwork}
Current approaches for video analysis exploit different amounts of spatio-temporal information in different ways depending on model architecture. 
One common strategy is to use models based on \textit{3D Convolutional Neural Networks (CNNs)} that use convolutions across temporal and spatial dimensions to model all local correlations~\cite{simonyan2014two,tran2014c3d} for supervised learning.
Similar architectures have been used for video prediction to directly generate the RGB values of pixels in future frames~\cite{mathieu2015deep,ranzato2014video,oh2015action,vondrick2016generating}.
Kalchbrenner et al.~\cite{kalchbrenner2016video} discussed that a general probabilistic model of videos should take into account the entire history (all context in past frames and generated pixels of present frame) for generating each new pixel.
However, their proposed Video Pixel Networks (VPNs) still use encoders based on stacks of convolution layers.
An inherent limitation of these models is that convolutions take only short-range dependencies into account due to the limited size of the kernels. 
These architectures need a larger stack of convolutional layers to use a wide context for reducing uncertainty. 
This increases the model capacity even though it may not be needed.

Recurrent neural networks are often used to address the issue of limited context.
Srivastava et al.\cite{srivastava2015unsupervised} proposed Long Short-Term Memory (LSTM)~\cite{Hochreiter:97lstm} based encoder-decoder models for the task of video prediction, but the canonical LSTM architecture used by them did not take the spatial structure of video data into account. 
This motivated the use of \textit{Convolutional LSTM (ConvLSTM)} based models which replace the internal transformations of an LSTM cell with convolutions.
Xingjian et al.~\cite{xingjian2015convolutional} proposed this design for precipitation nowcasting; the motivation being that the convolution operation would model spatial dependencies, while LSTM connectivity would offer increased temporal context.
The same modification of LSTM was simultaneously proposed by Stollenga et al.~\cite{stollenga2015parallel} for volumetric image segmentation under the name PyraMiD-LSTM, due to its close relationship with the Multi-Dimensional LSTM (MD-LSTM)~\cite{Graves2007}.

Recently, ConvLSTM has become a popular building block for video prediction models.
Finn et al.~\cite{finn2016unsupervised} used it to design a model that was trained to predict pixel motions instead of values.
Lotter et al.~\cite{lotter2016deep} developed the Deep Predictive Coding Network (PredNet) architecture inspired by predictive coding, which improves its own predictions for future frames by incorporating previous prediction errors.
It is also used in the MCNet~\cite{villegas2017decomposing} which learns to model the scene content and motion separately, and in the Dual Motion GAN~\cite{liang2017dual} which learns to produce consistent pixel and flow predictions simultaneously.
Wang et. al.~\cite{wang2017predrnn} have recently proposed the modification of stacked ConvLSTM networks for video prediction by sharing the hidden state among the layers in the stack.

For videos with mostly static backgrounds, it is helpful to explicitly model moving foreground objects separately from the background~\cite{simonyan2014two,vondrick2016generating,finn2016unsupervised}. 
Another active line of investigation is the development of architectures that only learn to estimate optical flow and use it to generate future frames instead of generating the pixels directly~\cite{patraucean2015spatio,liu2017video}.

Deterministic models trained with typical loss functions can result in imprecise predictions simply because the future is ambiguous given the past.
For example, if there are multiple possible future frames, models trained to minimize the L2 loss will generate their mean frame.
One approach for obtaining precise, natural-looking frame predictions in such cases is the use of adversarial training~\cite{mathieu2015deep,vondrick2016generating} based on Generative Adversarial Networks~\cite{goodfellow2014generative}.
Another is to use probabilistic models for modeling the distribution over future frames, from which consistent samples can be obtained without averaging of modes~\cite{xue2016visual,kalchbrenner2016video}. 

\section{Missing Contexts in Other Network Architectures}
\label{sec:context}
As mentioned earlier, blurry predictions can result from a video prediction model if it does not adequately capture all relevant information in the past video frames which can be used to reduce uncertainty.
\autoref{fig:convlstm} shows the recurrent connections of a pixel at time $t$ with a $3 \times 3$ convolution between two frames (left) and the information flow of a ConvLSTM predicting the pixel at time $T+1$ (right). 
The covering context grows progressively over time (depth), but there are also blind spots which cannot be used for prediction.
In fact, as can be seen in~\autoref{fig:convlstm} (right, marked in gray color), frames in the recent past have larger blind areas.
Due to this structural issue, the network is unable to capture the entire available context and is likely to miss important spatio-temporal dependencies leading to increased ambiguity in the predictions.
The prediction will eventually fail when the object appearance or motion in videos changes dramatically within a few frames. 

One possible way to address limited context, widely used in CNNs for image analysis, is to expand context by stacking multiple layers (sometimes with dilated convolutions~\cite{yu2015multi}).
However, stacking layers still limits the available context to a maximum as dictated by the network architecture, and the number of additional parameters required to gain sufficient context can be very large for high resolution videos.
Another technique that can help is using a multi-scale architecture, but fixed scale factors may not generalize to all possible objects, their positions and motions.

\section{Method}
\label{sec:method}

We introduce the Fully Context-aware Video Prediction model (ContextVP) --- an architecture that avoids blind spots by covering all the available context by design.
Its advantages are:
\begin{itemize}
  \item Since each processing layer covers the entire context, increasing depth is only used as necessary to add computation power, not more context. A priori specification of scale factors is also not required.
  \item Compared to models that utilize increased depth to cover larger context such as our baseline 20-layer models, more computations can be parallelized.
  \item Compared to state-of-the-art models from recent literature, it results in improved performance without the use of separation of motion and content, learning optical flow or adversarial training (although combinations with these strategies may further improve results).
\end{itemize}

Let $x_{1}^{T} = \{x_1, ...,x_{T}\}$ be a given input sequence of length $T$.
$x_{t} \in \mathds{R}^{H \times W \times C}$ is the $t$-th frame, where $t \in \{1, ..., T\}$, $H$ is the height, $W$ the width, and $C$ the number of channels. 
For simplicity, assume $C$ = 1, $x_{1}^{T}$ is then a cuboid of pixels bounded by six planes.
The task is to predict $p$ future frame(s) in the sequence, $x_{t+1}^{t+p} = \{x_{t+1}, ...,x_{t+p}\}$ (next-frame prediction if $p=1$).
Therefore, our goal is to integrate information from the entire cuboid $x_{1}^{T}$ into a representation at the plane where $t=T$, which can be used for predicting $x_{t+1}^{t+p}$.
This is achieved in the proposed model by using fully context-aware layers, each consisting of two blocks. 
The first block is composed of \textit{Parallel MD-LSTM units} that sequentially aggregate information from different directions. 
The second block is the \textit{Context Blending Block} that combines the output of PMD units for all directions. 
The context covered using PMD units for each direction (top) and the combined context from past frames (down) are visualized in~\autoref{fig:pyramid}.
The schematic in \autoref{fig:architecture} shows the overall architecture of our best model.

\subsection{Parallel MD-LSTM Unit}
Multidimensional LSTM (MD-LSTM)~\cite{Graves2007} networks, a specialization of DAG-RNNs \cite{baldi2003principled}, have been applied to various problems where the input is two-dimensional such as handwriting recognition~\cite{graves:2009nips}, 2D image classification~\cite{byeon2014texture} and segmentation~\cite{byeon15}. 
They consist of two MD-LSTM blocks per dimension to combine context from all possible directions.
In principle, MD-LSTM networks can be applied to any high-dimensional domain problem (including video prediction) to model all available dependencies in the data compactly.
However, the fully sequential nature of the model makes it unsuitable for parallelization and thus impractical for higher dimensional data. 
The \textit{PyraMiD-LSTM}~\cite{stollenga2015parallel} addressed this issue by re-arranging the recurrent connection topology of each MD-LSTM block from cuboid to pyramidal (for 3D data).
It could be implemented efficiently by utilizing the convolution operation. 
So far, the idea of using LSTM to aggregate information from all directions was only explored in a limited setting (2D/3D image segmentation).

\begin{figure}[t]
    \centering
        \includegraphics[width=0.8\textwidth]{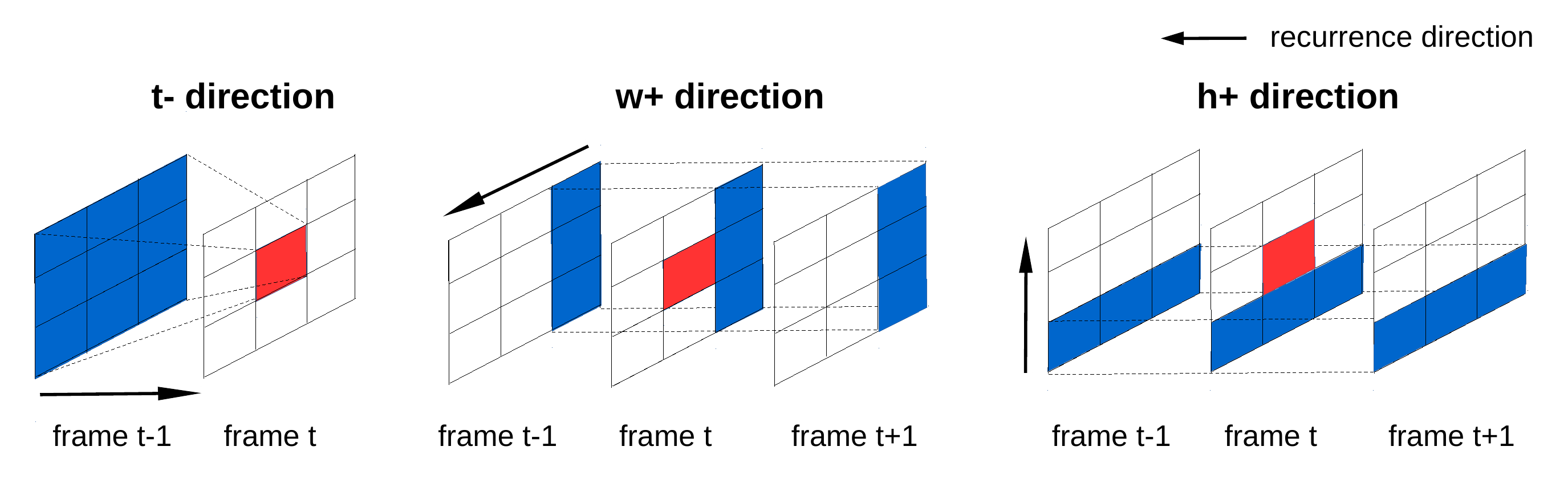}   
\caption{
Illustration of one computation step of PMD units for $t-$, $w+$, and $h+$ recurrence directions.
Each unit computes the activation at the current position (red) using context from a fixed receptive field (here 3$\times$3) of the previous frame along its recurrence direction (blue).
This computation is efficiently implemented using convolutions.
}
\label{fig:computations}
\end{figure}

We refer to the parallel computing units used in the PyraMiD-LSTM architecture simply as Parallel Multi-Dimensional (PMD) units since they model contextual dependencies in a way that is amenable to parallelization.
They are mathematically similar to ConvLSTM units but our terminology highlights that it is \textbf{not} necessary to limit convolutional operations to spatial dimensions and LSTM connectivity to the temporal dimension as is conventional. 
As can be seen in \autoref{fig:computations}, PMD units can be used to aggregate context along any of the six directions available in a cuboid. 
Three directions are shown: $t\sminus$, $w\splus$, and $h\splus$.
At each plane, the local computation for each pixel is independent of other pixels in the same plane, so all pixels are processed as parallel using the convolution operation.
The computational dependencies across planes are modeled using the LSTM operation.
Computations for each PMD unit are explained mathematically below.

For any sequence of $K$ two dimensional planes $x_1^K = \{x_1, ..., x_K\}$, the PMD unit computes the current cell and hidden state $c_k, s_k$ using input, forget, output gates $i_k, f_k, o_k$, and the transformed cell $\tilde{c}_k$ given the cell and hidden state from the previous plane, $c_{k-1}, s_{k-1}$.
\begin{equation}
\begin{aligned}
    i_k &= \sigma(W_{i} \ast x_k + H_{i}  \ast s_{k \sminus 1}  + b_{i}), \\
    f_k &= \sigma(W_{f} \ast x_k + H_{f}  \ast s_{k \sminus 1} + b_{f}), \\ 
    o_k &= \sigma(W_{o} \ast x_k + H_{o}  \ast s_{k \sminus 1}  + b_{o})\\ 
    \tilde{c}_k &= \tanh(W_{\tilde{c}} \ast x_k + H_{\tilde{c}}  \ast s_{k \sminus 1} + b_{\tilde{c}}), \\ 
    c_k &=  f_k \odot c_{k \sminus 1} + i_k \odot \tilde{c}_k, \\
    s_k &= o_k \odot \tanh(c_k).
\end{aligned}
\end{equation}
Here ($\ast$) is the convolution operation, and ($\odot$) the element-wise multiplication. 
$W$ and $H$ are the weights for input-state and state-state. 
The size of weight matrices are dependent only on the kernel size and number of units. 
If the kernel size is larger, more local context is taken into account. 

As shown in \autoref{sec:context}, using a ConvLSTM would be equivalent to running a PMD unit along the time dimension from $k=1$ to $k=T$, which would only integrate information from a pyramid shaped region of the cuboid and ignore several blind areas.
For this reason, it is necessary to use four additional PMD units, for which the conditioning directions are aligned with the spatial dimensions, as shown in \autoref{fig:pyramid} (top).
We define the resulting set of five outputs at frame $T$ as ${s^d}$ where $d \in D = \{h\sminus, h\splus, w\sminus, w\splus, t\sminus \}$ denotes the recurrence direction.
Together this set constitutes a representation of the cuboid of interest $x_1^T$.
Outputs at other frames in $x_1^{T-1}$ are ignored.

\subsection{Context Blending Block} 
This block captures the entire available context by combining the output of PMD units from all directions at frame $T$. 
This results in the critical difference from the traditional ConvLSTM: the context directions are aligned not only with the time dimension but also with the spatial dimensions. 
We consider two ways to combine the information from different directions.

\textbf{Uniform blending (U-blending):} this strategy was used in the traditional MD-LSTM~\cite{graves:icann2007,byeon15} and PyraMiD LSTM~\cite{stollenga2015parallel}. 
It simply sums the output of all directions along the channel dimension and then applies a non-linear transformation on the result: 
 \begin{equation}
 m = f((\sum_{d \in D} s^{d}) \cdot W + b), 
 \label{eq:blending1}
\end{equation}
where $W \in \mathds{R}^{N1 \times N2}$ and $b \in \mathds{R}^{N2}$ are a weight matrix and a bias. 
$N1$ is the number of PMD units, and $N2$ is the number of (blending) blocks.
$f$ is an activation function.

\textbf{Weighted blending (W-blending):} 
the summation of PMD unit outputs in U-blending assumes that the information from each direction is equally important for each pixel.
We propose W-blending to remove this assumption and learn the relative importance of each direction during training with the addition of a small number of additional weights compared to the overall model size.
The block concatenates $s$ from all directions:
\begin{equation}
S = 
\begin{bmatrix}
s^{t-} &
s^{h-} &
s^{h+} &
s^{w-} &
s^{w+}\\
\end{bmatrix}^{T}
\end{equation}
The vector $S$ is then weighted as follows: 
\begin{equation}
m = f(S \cdot W + b), 
\label{eq:blending2}
\end{equation}
where $W \in \mathds{R}^{(5 \times N1) \times N2}$ ($5$ is the number of directions). 
\autoref{eq:blending1} and \autoref{eq:blending2} 
are implemented using $1\times1$ convolutions. 
We found that W-blending is crucial for achieving high performance for the task of video prediction (see \autoref{tab:ablation}).

\subsection{Directional Weight-Sharing (DWS)}
Visual data tend to have structurally similar local patterns along opposite directions.
This is the reason why horizontal flipping is a commonly used data augmentation technique in computer vision.
We propose the use of a similarly inspired weight-sharing technique for regularizing the proposed networks.
The weights and biases of the PMD units in opposite directions are shared i.e. weights for $h\sminus$ and $h\splus$ are shared, as are $w\sminus$ and $w\splus$. 
This strategy has several benefits in practice: 1) it lowers the number of parameters to be learned, 2) it incorporates knowledge about structural similarity into the model, and 3) it improves generalization. 

\subsection{Training}
$\hat{x} = g(m)$ is an output of the top-most (output) layer, where $g$ is an output activation function. 
The model minimizes the loss between the predicted pixels and the target pixels. 
$\mathcal{L}_p$ loss and the Image Gradient Difference Loss (GDL)~\cite{mathieu2015deep} are combined. 
By keeping the loss function simple, the results reflect the impact of having access to all available context.
Let $y$ and $\hat{x}$ be the target and the predicted frame. 
The objective function is defined as follows:
\begin{equation}
\small
\begin{aligned}
\mathcal{L}(y, \hat{x}) &= \lambda_{p} \mathcal{L}_{p} (y, \hat{x}) + \lambda_{gdl} \mathcal{L}_{gdl} (y, \hat{x}) \\
\mathcal{L}_{p} (y, \hat{x}) &=  ||y - \hat{x}||_{p} \\
\mathcal{L}_{gdl} (y, \hat{x}) &= \sum_{i,j} |y_{i,j} - y_{i-1,j}| - |\hat{x}_{i,j} - \hat{x}_{i-1,j}| + |y_{i,j-1} - y_{i,j}| - |\hat{x}_{i,j-1} - \hat{x}_{i,j}|,
\end{aligned}
\label{eq:loss}
\end{equation}
where $|.|$ is the absolute value function, $\hat{x}_{i,j}$ and $y_{i,j}$ are the pixel elements from the frame $\hat{x}$ and $y$, respectively. 
$\lambda_{p}$ and $\lambda_{gdl}$ are the weights for each loss. In our experiments, $\lambda_{gdl}$ is set to $1$ when $p=1$ and $0$ when $p=2$. $\lambda_{p}$ is always set to $1$. 

We use ADAM~\cite{kingma2014adam} as the optimizer with an initial learning rate of $1e-3$. The learning rate is decayed every $5$ epochs with the decay rate $0.99$.
Weights are initialized using the Xavier's normalized initializer~\cite{glorot2010understanding} 
and the states of the LSTMs are initialized to zero (no prior). 

\section{Experiments}
\label{sec:experiment}
We evaluate the proposed approach on three real-world scenarios with distinct characteristics: human motion prediction (Human 3.6M dataset~\cite{ionescu2014human3}), car-mounted camera video prediction (train: KITTI dataset~\cite{geiger2013vision}, test: CalTech Pedestrian dataset~\cite{dollar2009pedestrian}, and human activity prediction (UCF-101 dataset~\cite{soomro2012ucf101}). 
All input pixel values are normalized to the range $[0, 1]$.
For human motion and car-mounted videos, the models are trained to use ten frames as input for predicting the next frame.
For the UCF-101 dataset, the input consists of four frames for fair comparison to past work.
Quantitative evaluation on the test sets is performed based on mean Peak Signal-to-Noise Ratio (PSNR) and the Structural Similarity Index Measure (SSIM)~\cite{wang2004image}\footnote{Mean Squared Error (MSE) is also reported for the car-mounted camera video prediction to compare with PredNet~\cite{lotter2016deep}.}.
These commonly used numerical measures are known to be not fully representative of human vision. 
Therefore, we highly recommend looking at the visual results in \autoref{fig:result-caltech} and \autoref{fig:result-ucf}. 

\begin{figure}[t]
    \centering
        \includegraphics[width=0.8\textwidth]{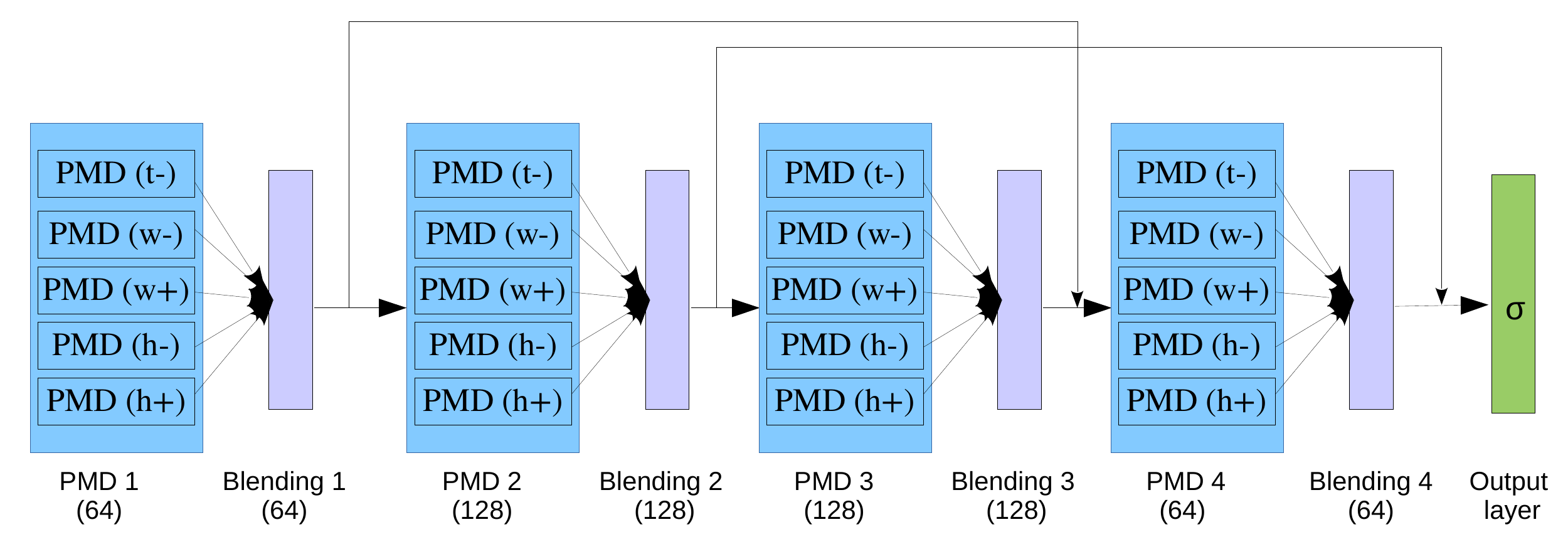}   
\caption{ContextVP-big (4 layers) architecture: Each layer contains 5 PMD units followed by context blending block. Two skip connections are used, which simply concatenate outputs of two layers (layers 1 \& 3, and 2 \& 4). The output layer uses the sigmoid activation function which outputs values in the range $(0, 1)$. ($\cdot$) indicates the number of hidden units in each layer. The ContextVP-small architecture has half the hidden units at each layer.}
\label{fig:architecture}
\vspace{-0.5cm}
\end{figure}

\vspace{0.2cm}
\textbf{Network architecture: } 
our best model architecture is illustrated in~\autoref{fig:architecture}.
It consists of a stack of four context-aware layers with skip connections that directly predicts the scaled RGB values of the next frame.
All results are reported for models using $3 \times 3$ convolutional kernels for all PMD units, identity activation function in \autoref{eq:blending1} and \autoref{eq:blending2} 
and training using $\mathcal{L}_1$ ($p=1$ in~\autoref{eq:loss}) with GDL loss.
Changing to $5 \times 5$ size kernels, use of nonlinear activations (e.g., ReLU~\cite{nair2010rectified} or $\mathrm{tanh}$) or layer normalization~\cite{ba2016layer} in the blending blocks does not affect the performance in our experiments. 
Finn et al.\cite{finn2016unsupervised} reported that $\mathcal{L}_1$ with the GDL loss function performs better than $\mathcal{L}_2$ but their performance in our case was very similar. 

\vspace{0.2cm}
\textbf{Baseline:}
our baseline (ConvLSTM20) is a network consisting of a stack of 20 ConvLSTM layers with kernels of size $3 \times 3$.
The number of layers was chosen to be 20 to cover a large context and also since each layer in our 4-layer model consists of 5 PMD units. 
Two skip connections similar to our model were also used. 
The layer sizes are choen to keep the number parameters comparable to our best model (ContextVP4-WD-big).
Surprisingly, \textbf{this baseline outperforms almost all state of the art models} except Deep Voxel Flow~\cite{liu2017video} on the UCF-101 dataset.
Note that it is \textbf{less amenable to parallelization} compared to ContextVP models where PMD units for different directions can be applied in parallel.

\begin{table}[t]
   \centering
   \footnotesize
      \caption{
      Results of ablation study on the Human3.6M dataset. 
      The model is trained on 10 frames and predicts the next frame. 
      The results are averaged over test videos.  
      D indicates directional weight-sharing, and U and W indicate uniform and weighted blending, respectively.
      Higher values of PSNR/SSIM and lower values of MSE indicate better results.
      }
      \begin{tabular}{c @{\hspace{0em}} c @{\hspace{0.9em}} c @{\hspace{1em}} c @{\hspace{0.9em}} c @{\hspace{0.9em}} c @{\hspace{1em}} c}
      \toprule
      name & \# layers & blending type & DWS & PSNR  & SSIM & \# parameters\\
      \midrule
      ContextVP1 & 1 & uniform (U) & N & 38.1  & 0.990 &  0.7M \\
      ContextVP3 & 3 & uniform (U) & N & 41.2  & 0.992 &  1.6M \\
      ContextVP4-U-big & 4 & uniform (U) & N  & 42.3    & 0.994 &   14.0M \\
      ContextVP4-W-big & 4 & weighted (W) & N  & 44.8   & 0.996 &  14.2M \\
      ContextVP4-WD-small & 4 & weighted (W) & Y  & 45.0   & 0.996 & 2.0M \\
      ContextVP4-WD-big & 4 & weighted (W) & Y  & \textbf{45.2}   & \textbf{0.996} & 8.6M \\
      \bottomrule
      \end{tabular}
      \label{tab:ablation}
      \vspace{-0.6cm}
\end{table}

\begin{table}[t]
   \centering
   \footnotesize
      \caption{Evaluation of Next-Frame Predictions on the Human3.6M dataset. 
      All models are trained on 10 frames and predicts the next frame. 
      The results are averaged over test videos. 
      ConvLSTM20 is our baseline containing 20 ConvLSTM layers. 
      Higher values of PSNR and SSIM, lower values of MSE indicate better results.
      Our best models (ContextVP4-WD: 4 layers with weighted blending and DWS) outperform our baseline as well as current state-of-the-art methods with fewer number of parameters.
      }
      \vspace{0.2cm}
      \begin{tabular}{l @{\hspace{0.9em}} c @{\hspace{0.9em}} c @{\hspace{0.9em}} c @{\hspace{0.9em}} c}
      \toprule
      Method  & PSNR  & SSIM & \#parameters & time (s)\\
      \midrule
      Copy-Last-Frame & 32  & - & - & -\\ 
      \midrule
      BeyondMSE~\cite{mathieu2015deep} & 26.7 & - & 8.9M & - \\
      PredNet~\cite{lotter2016deep} & 38.9 & - & 6.9M & -\\
      \midrule
      ConvLSTM20 &  44.1 &  0.995 & 9.0M &  0.153 \\
      \midrule
      ContextVP4-WD-small  & 45.0   & 0.996 & 2.0M & - \\
      ContextVP4-WD-big & \textbf{45.2}   & \textbf{0.996}& 8.6M & \textbf{0.092} \\
      \bottomrule
      \end{tabular}
      \vspace{-0.3cm}
      \label{tab:comparison-human}
\end{table}
\subsection{Human Motion Prediction (Human3.6M dataset)}
We first evaluate our model on Human3.6M dataset~\cite{h36m_pami}. 
The dataset includes seven human subjects (three females and four males). 
Five subjects are used for training and the other two for validation and testing. 
The videos are subsampled to 10 fps and downsampled to $64 \times 64$ resolution. 

\vspace{0.2cm}
\textbf{Ablation study:} 
using this dataset, we evaluate the importance of various components of our model: multiple layers, types of context blending, and DWS regularization.
\autoref{tab:ablation} shows the results.
We find that performance increases substantially with number of layers, switch to W-blending, and addition of DWS.
ContextVP1, ContextVP3 and ContextVP4-U-big use U-blending and no DWS, corresponding to direct adaptation of PyraMiD-LSTM for video prediction.

\vspace{0.2cm}
\textbf{Comparison to other methods:} 
\autoref{tab:comparison-human} shows the comparison of the prediction results with the Baseline ConvLSTM as well as PredNet~\cite{lotter2016deep}, and BeyondMSE~\cite{mathieu2015deep}. 
Another baseline Copy-Last-Frame is included to show the result of simply copying the last input frame. 
We do not compare to Finn et al.~\cite{finn2016unsupervised} since their model was not trained for next-frame prediction.
From \autoref{tab:ablation}, it can be seen that single layer ContextVP already outperforms BeyondMSE which uses 3D-CNN, and the three-layer ContextVP networks outperform PredNet which uses ConvLSTM. 
Finally, four layer ContextVP networks with W-blending and DWS outperform all approaches, even with much fewer parameters (ContextVP4-WD-small).
Increasing the model size (ContextVP4-WD-big) brings a minor improvement in final performance.

\noindent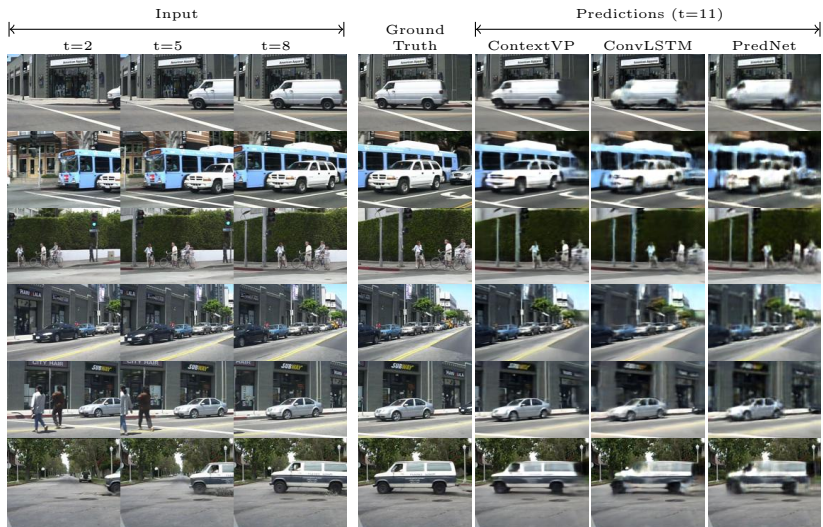
\begin{figure*}[t]
\centering
\tiny
\begin{tabular}{  c @{\hspace{0.7em}} c @{\hspace{0.2em}} c @{\hspace{0.2em}} c @{\hspace{0.2em}} c }
    Input &  & \multicolumn{3}{c}{Predictions (t=11)} \\
      & Ground & \tikzmark{a} & & \tikzmark{b}\\
     t=2 \quad \quad \quad t=5 \quad \quad \quad  \quad t=8  & Truth & ContextVP & ConvLSTM & PredNet \\

    \includegraphics[width=4.5cm]{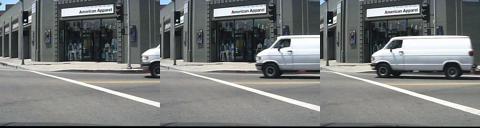} &
    \includegraphics[width=1.5cm]{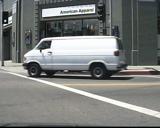} &
    \includegraphics[width=1.5cm]{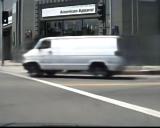} &
    \includegraphics[width=1.5cm]{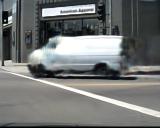} &
    \includegraphics[width=1.5cm]{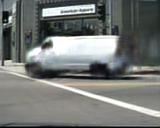} \\ [-1em]

     \includegraphics[width=4.5cm]{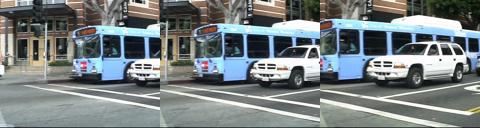} &
    \includegraphics[width=1.5cm]{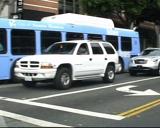} &
    \includegraphics[width=1.5cm]{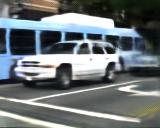} &
    \includegraphics[width=1.5cm]{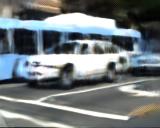} &
    \includegraphics[width=1.5cm]{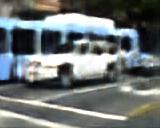} \\ [-1em]
       
    \includegraphics[width=4.5cm]{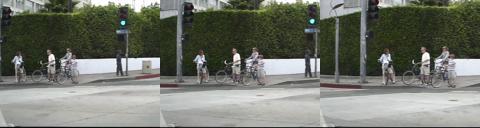} &
    \includegraphics[width=1.5cm]{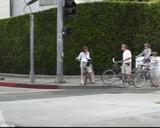} &
    \includegraphics[width=1.5cm]{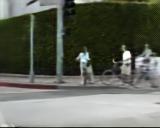} & 
    \includegraphics[width=1.5cm]{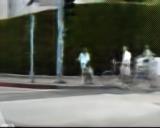} &
    \includegraphics[width=1.5cm]{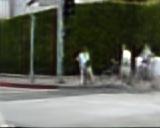} \\ [-1em]

    \includegraphics[width=4.5cm]{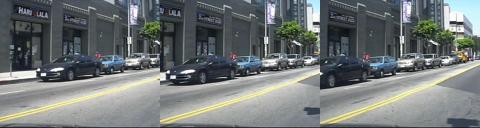} &
    \includegraphics[width=1.5cm]{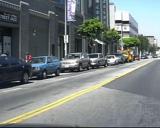} &
    \includegraphics[width=1.5cm]{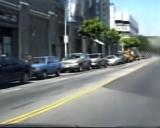} &
    \includegraphics[width=1.5cm]{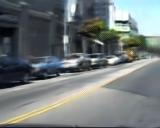} &
    \includegraphics[width=1.5cm]{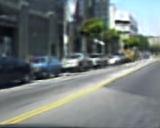} \\ [-1em]

    \includegraphics[width=4.5cm]{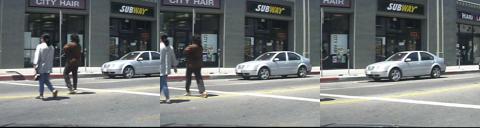} &
    \includegraphics[width=1.5cm]{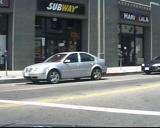} &
    \includegraphics[width=1.5cm]{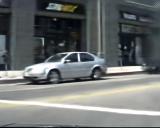} &
    \includegraphics[width=1.5cm]{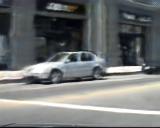} &
    \includegraphics[width=1.5cm]{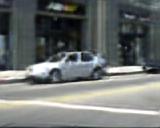} \\ [-1em]

    \includegraphics[width=4.5cm]{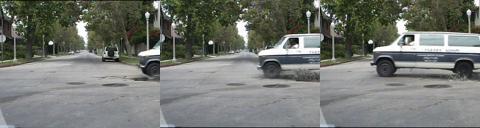} &
    \includegraphics[width=1.5cm]{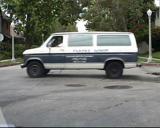} &
    \includegraphics[width=1.5cm]{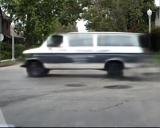} &
    \includegraphics[width=1.5cm]{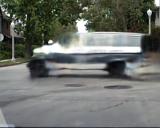} &
    \includegraphics[width=1.5cm]{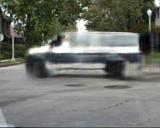} \\ 
  
\end{tabular}
\begin{tikzpicture}[overlay, remember picture, yshift=.25\baselineskip, shorten >=.5pt, shorten <=.5pt,align=left]
    \draw[|<->|] (-10.95,3.2) -- (-6.45,3.2);
    \draw[|<->|] (-4.75,3.2) -- (-0.12,3.2);
  \end{tikzpicture}
\caption{Qualitative comparisons from the test set among our best model (ContextVP4-WD-big), the baseline (ConvLSTM20), and the state-of-the-art model (PredNet). All models are trained for next-frame prediction given 10 input frames on the KITTI dataset, and tested on the CalTech Pedestrian dataset.}
\label{fig:result-caltech}
\vspace{-0.5cm}
\end{figure*}

\begin{table}[t]
\centering
\footnotesize
\caption{
Evaluation of Next frame prediction on the CalTech Pedestrian dataset (trained on the KITTI dataset). 
All models are trained on 10 frames and predicts the next frame. 
The results are averaged over test videos. 
ConvLSTM20 is our baseline containing 20 ConvLSTM layers. 
Higher values of PSNR and SSIM, lower values of MSE indicate better results.
(\textbf{+}) This score is provided by~\cite{liang2017dual}. 
(\textbf{*}) The scores provided in Lotter et al.~\cite{lotter2016deep} are averaged over nine frames (time steps 2--10 in their study), but ours are computed only on the next predicted frame. 
We therefore re-calculated the scores of PredNet using their trained network. 
Our best models (ContextVP4-WD: 4 layers with weighted blending and DWS) outperform the baseline as well as current state-of-the-art methods with fewer number of parameters.
}
\vspace{0.2cm}
\begin{tabular}{l @{\hspace{0.9em}} c @{\hspace{0.9em}} c @{\hspace{0.9em}} c @{\hspace{0.9em}} c @{\hspace{0.9em}} c}
\toprule
{Method}  & MSE ($\times 10^{-3}$) & {PSNR} & {SSIM} & {\#parameters} & time (s)\\
\midrule
Copy-Last-Frame & 7.95  & 23.3  & 0.779 & - & -\\ 
\midrule
\textbf{+}BeyondMSE~\cite{mathieu2015deep} & 3.26 & - & 0.881 & - & - \\
\textbf{*}PredNet~\cite{lotter2016deep} & 2.42 & 27.6  & 0.905 & 6.9M & -\\
Dual Motion GAN~\cite{liang2017dual} & 2.41 & - & 0.899 & 113M & -\\
\midrule
ConvLSTM20 & 2.26 & 28.0 & 0.913 &  9.0M & 0.447 \\
\midrule
ContextVP4-WD-small & 2.11     & 28.2   & 0.912 & 2.0M & -\\
ContextVP4-WD-big & \textbf{1.94}     & \textbf{28.7}   & \textbf{0.921} & 8.6M & \textbf{0.346} \\
\bottomrule
\end{tabular}
\vspace{-0.3cm}
\label{tab:result-caltech}
\end{table}

\subsection{Car-mounted Camera Video Prediction (KITTI and CalTech Pedestrian dataset)}
The model is trained on the KITTI dataset~\cite{geiger2013vision} and tested on the CalTech Pedestrian dataset~\cite{dollar2009pedestrian}.
Every ten input frames from ``City'', ``Residential'', and ``Road'' videos are sampled for training resulting in $\approx$41K frames. 
Frames from both datasets are center-cropped and down-sampled to  $128 \times 160$ pixels. 
We use the exact data preparation as PredNet~\cite{lotter2016deep} for direct comparison. 

The car-mounted camera videos are taken from moving vehicles and consist of a wide range of motions. 
Compared to Human3.6M, which has static background and small motion flow, this dataset has diverse and large motion of cars at different scales and also has large camera movements. 
To make predictions for such videos, a model is required to learn not only small movement of pedestrians, but also relatively large motion of surrounding vehicles and backgrounds. 

We compare our approach with the Copy-Last-Frame and ConvLSTM20 baselines as well as BeyondMSE, PredNet, and Dual Motion GAN~\cite{liang2017dual} which are the current best models for this dataset. 
Note that the scores provided in Lotter et al.~\cite{lotter2016deep} are averaged over nine frames (time steps 2--10 in their study), but ours are computed only on the next predicted frame. 
Therefore, we re-calculated the scores of PredNet for the next frame using their trained network. 
As shown in \autoref{tab:result-caltech}, our four layer model with W-blending and DWS outperforms the state-of-the-art on all metrics.
Once again, the smaller ContextVP network already matches the baseline while being much smaller and more suitable for parallelization. 
Some samples of the prediction results from the test set are provided in~\autoref{fig:result-caltech}. 
Our model is able to adapt predictions to the current scene and make sharper predictions compared to the baseline and PredNet.  

\begin{table}[t]
    \centering
   \footnotesize
    \caption{Evaluation of Next-Frame Predictions on the UCF-101 dataset. Models are trained on four frames and predict the next frame. 
    Results are averaged over test videos. 
    ConvLSTM20 is our baseline containing 20 ConvLSTM layers. 
    (\textbf{*}) Liu et al.~\cite{liu2017video} did not provide the number of parameters but noted that their model has the same number of parameters as BeyondMSE~\cite{mathieu2015deep}. 
    Higher values of PSNR and SSIM, lower values of MSE indicate better results.
    For UCF-101 dataset, larger kernel size produces the better prediction using fewer number of parameters.
    Our best models (ContextVP4-WD: 4 layers with weighted blending and DWS) outperform the baseline as well as current state-of-the-art methods with fewer number of parameters.
    }
    \vspace{0.2cm}
    \begin{tabular}{l @{\hspace{0.9em}} c  @{\hspace{0.9em}}c  @{\hspace{0.9em}}c @{\hspace{0.9em}}c}
    \toprule
    Method  & PSNR  & SSIM & \# parameters & time (s) \\
    \midrule
    BeyondMSE~\cite{mathieu2015deep} & 32 & 0.92 & 8.9M &-\\
    MCnet+RES~\cite{villegas2017decomposing} &  31 & 0.91 & 14M & -\\
    DVF~\cite{liu2017video}  & 33.4 & \textbf{0.94} & $\approx$8.9M\textbf{*} & -\\
    \midrule
    ConvLSTM20 & 32.9 & 0.91 & 9.0M & 0.499 \\
    \midrule
    ContextVP4-WD-small & 34.7   & 0.92 & 2M & - \\
    ContextVP4-WD-big  & \textbf{34.9}  & 0.92 & 8.6M & \textbf{0.474}\\
    \bottomrule
    \end{tabular}
    \label{tab:result-ucf}
    \vspace{-0.4cm}
\end{table}

\subsection{Human Action Prediction (UCF-101 dataset)}
The last dataset we test on is UCF-101~\cite{soomro2012ucf101} consisting of videos from  YouTube. 
Although many videos in this dataset contain small movements between frames, they contain much more diversity in objects, backgrounds and camera motions compared to previous datasets. 
Our experimental setup follows that of Mathieu et al.~\cite{mathieu2015deep}.
About 500K training videos are selected from the UCF-101 training set, and $10\%$ of UCF-101 test set is used for testing (378 videos). 
All frames are resized to $256 \times 256$. 
Note that Mathieu et al. used randomly selected sequences of $32 \times 32$ patches from the Sports-1M dataset~\cite{KarpathyCVPR14} for training since the motion between frames in the UCF-101 dataset are too small to learn dynamics.
Our model however, is directly trained on UCF-101 subsequences of length four with the original resolution. 
Motion masks generated using Epicflow~\cite{revaud:hal-01142656} provided by Mathieu et al. are used for validation and testing, so the evaluation is focused on regions with significant motion when computing PSNR and SSIM.

\autoref{tab:result-ucf} presents the quantitative comparison to the baseline as well as four best results from past work: adversarial training (BeyondMSE; \cite{mathieu2015deep}), the best model from Villegas et al. (MCnet+RES; \cite{villegas2017decomposing}), and Deep Voxel Flow (DVF; \cite{liu2017video}).
The results are similar to previous datasets: even at much smaller size, the four layer ContextVP network outperforms the baseline and other methods, and increasing the model size brings a small improvement in score. 
However, it does not outperform DVF on SSIM score.
These results shows that small ContextVP models can capture relevant spatial-temporal information and use it to make sharp predictions in very diverse settings without requiring adversarial training. 

\vspace{-0.2cm}
\begin{figure}[t]
    \centering
    \begin{subfigure}{0.6\textwidth}
         \includegraphics[trim={0 0 0.3cm 0.5cm},clip, width=\textwidth]{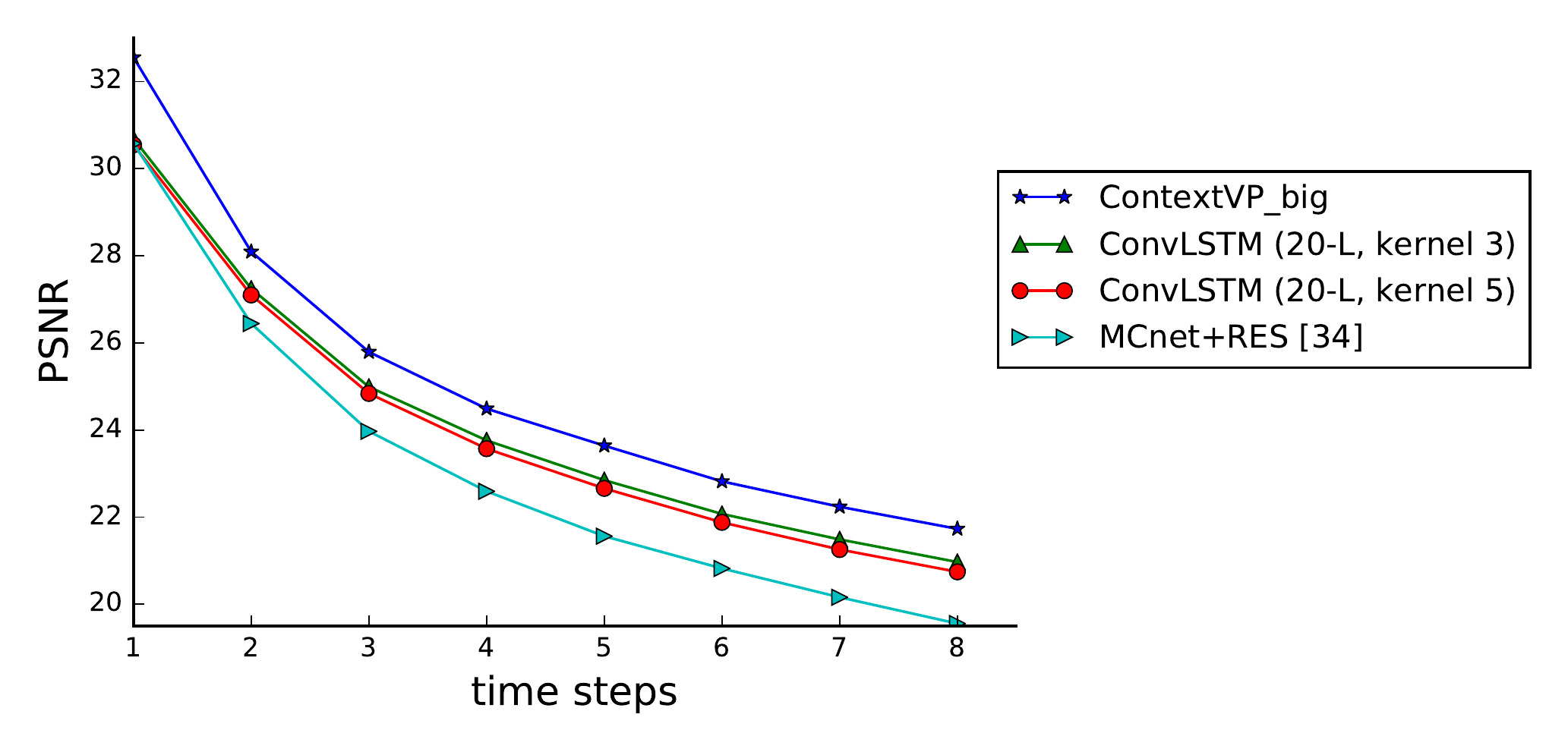}
    \end{subfigure}
    \begin{subfigure}{0.39\textwidth}
         \includegraphics[trim={0.5cm 0 0 0},clip, width=\textwidth]{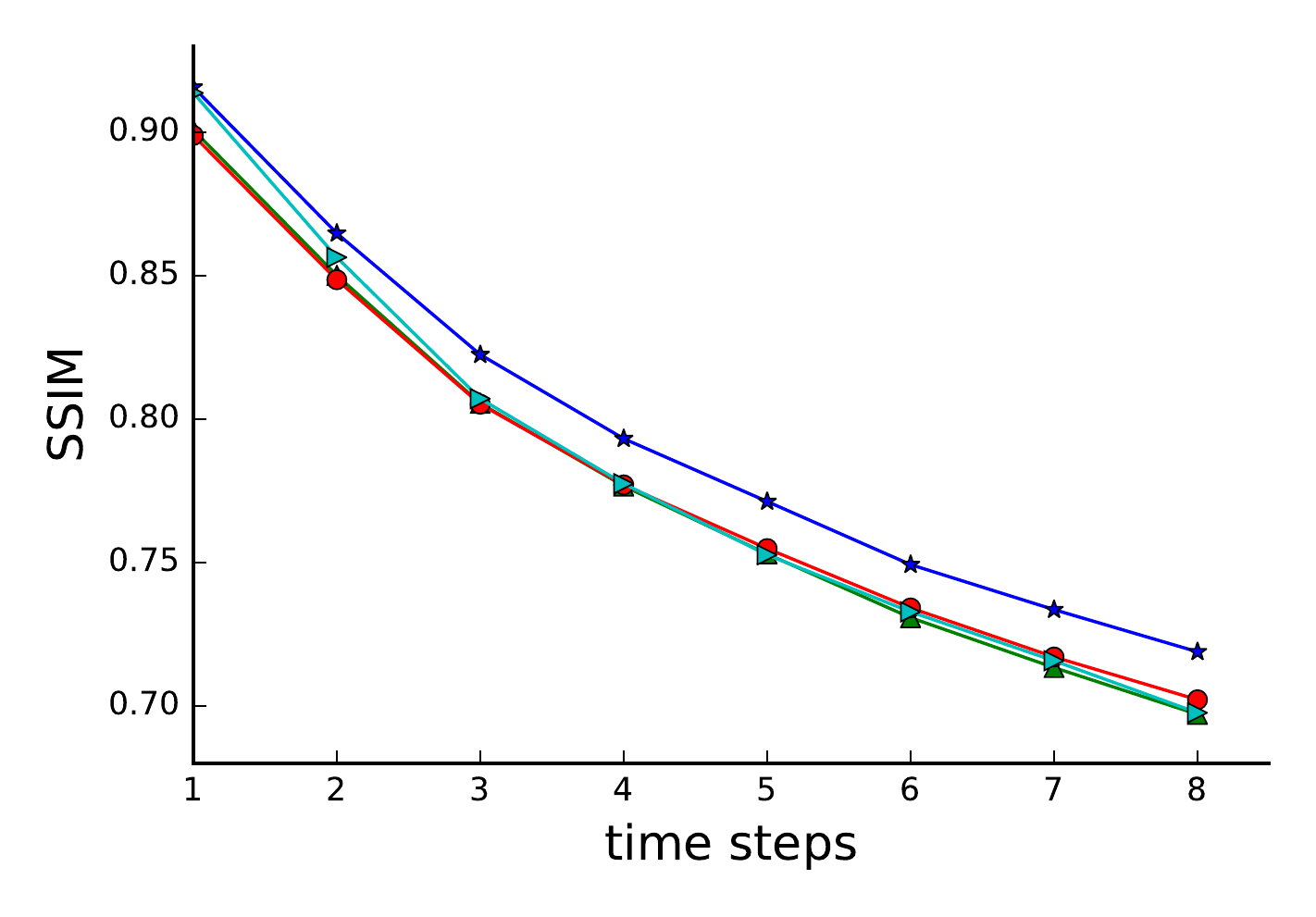}
    \end{subfigure}
    \caption{Comparison of multi-step prediction on UCF101: our best models (ContextVP), Villegas et al. (MCnet+RES; \cite{villegas2017decomposing}), and 20-layer ConvLSTM baseline. 
    Given 4 input frames, the models are trained for next-frame prediction and tested to predict 8 frames recursively.}
    \label{fig:comparison_mcnet}
    \vspace{-0.5cm}
\end{figure}

\vspace{0.4cm}
\textbf{Multi-Step Prediction:} 
\autoref{fig:comparison_mcnet} compares multi-step prediction results of our models with the baseline (ConvLSTM20), MCnet+RES, and BeyondMSE. 
Given four frames, all networks were trained for single frame prediction and scored on the test set by predicting eight frame recursively. 
Our small and big models perform very similarly according to PSNR, but the SSIM score for further predictions are better for the \emph{smaller} model.
Qualitative comparisons are presented in \autoref{fig:result-ucf}.
In the first video, ContextVP produces clear predictions for the subject's face and fewer motion artifacts for the black object, as opposed to other methods.
In the second video, more details of the rider and the horse are preserved by ContextVP.

\noindent\begin{figure*}[t]
\centering
\tiny
\begin{tabular}{  c @{\hspace{0.7em}} c }
    Input & Ground Truth and Predictions\\
     & \\ 
    t=2 \quad \quad \quad t=4 &  t=5 \quad \quad \quad t=7 \quad \quad \quad t=9 \quad \quad \quad t=11 \\
    \includegraphics[width=2.6cm]{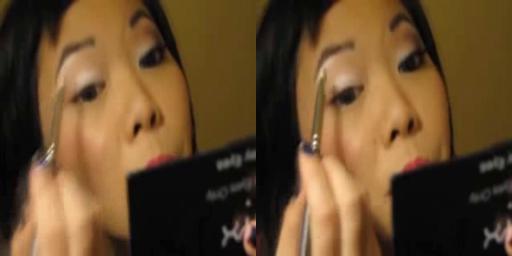} &
    \includegraphics[width=5.2cm]{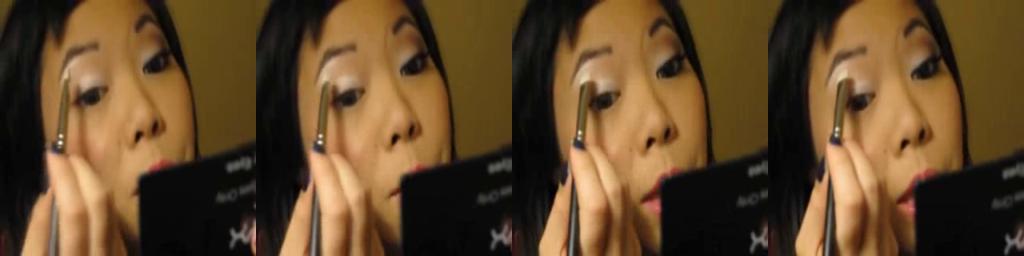} \\
    \quad \quad ContextVP & \includegraphics[align=c, width=5.2cm]{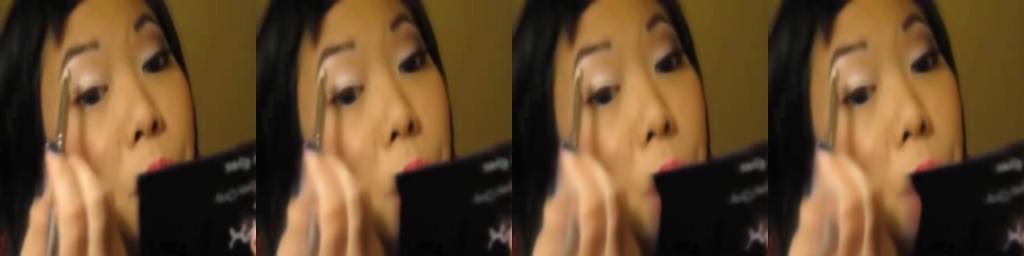} \\
    \quad \quad ConvLSTM & \includegraphics[align=c, width=5.2cm]{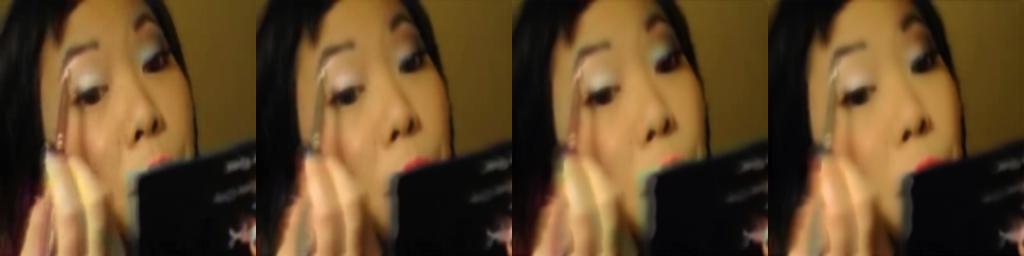} \\
    \quad \quad MCnet & \includegraphics[align=c, width=5.2cm]{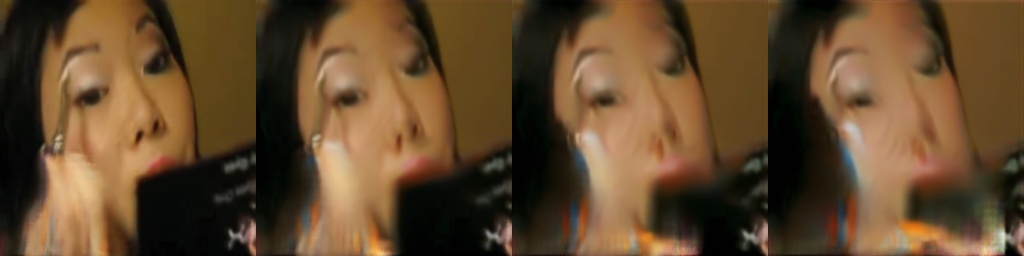} \\  [-1em]

    \includegraphics[width=2.6cm]{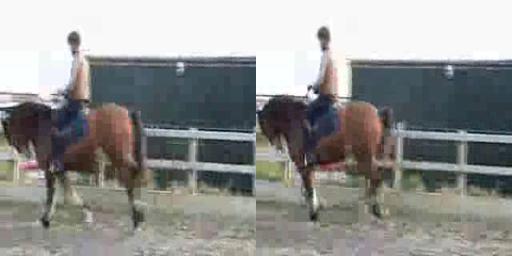} &
    \includegraphics[width=5.2cm]{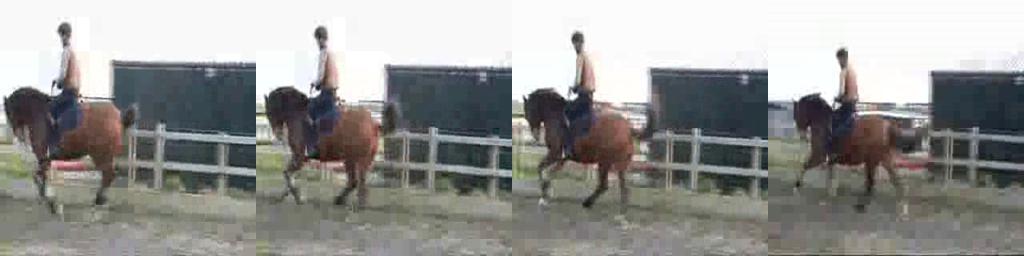} \\
    \quad \quad ContextVP & \includegraphics[align=c, width=5.2cm]{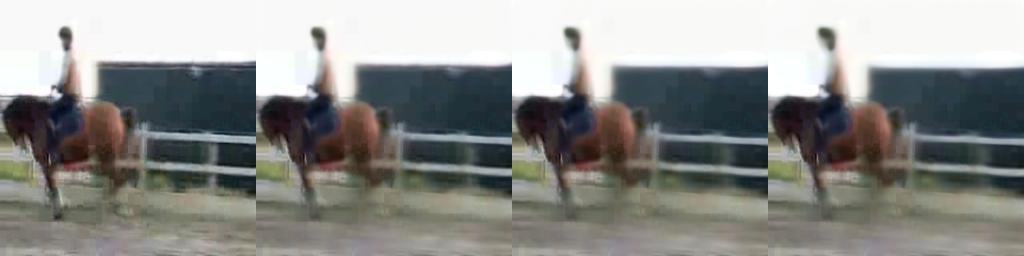} \\
    \quad \quad ConvLSTM & \includegraphics[align=c, width=5.2cm]{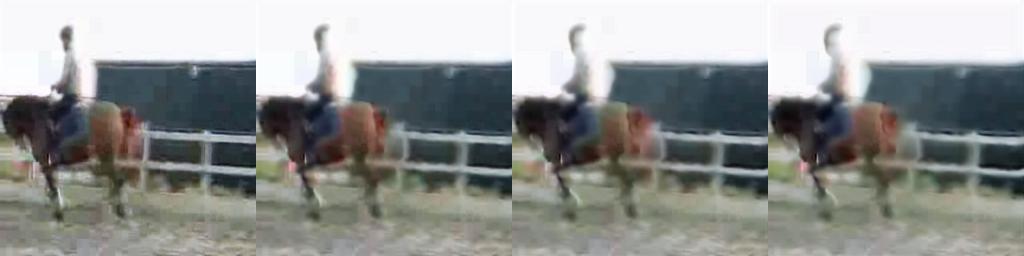} \\
    \quad \quad MCnet & \includegraphics[align=c, width=5.2cm]{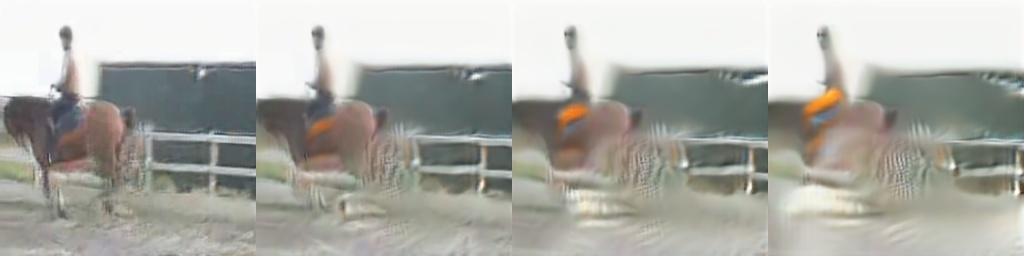} \\ 

\end{tabular}
\begin{tikzpicture}[overlay, remember picture, yshift=.25\baselineskip, shorten >=.5pt, shorten <=.5pt,align=left]
    \draw[|<->|] (-8.1,5.1) -- (-5.5,5.1);
    \draw[|<->|] (-5.4,5.1) -- (-0.1,5.1);
  \end{tikzpicture}
\vspace{.2cm}
\caption{Qualitative comparisons from the UCF-101 test set among our best model (ContextVP4-WD-big), the baseline (ConvLSTM20), and the state-of-the-art model (MCNet). All models are trained for next-frame prediction given 4 input frames. They are then tested to recursively predict 8 future frames (see also \autoref{fig:comparison_mcnet}).}
\label{fig:result-ucf}
\end{figure*}
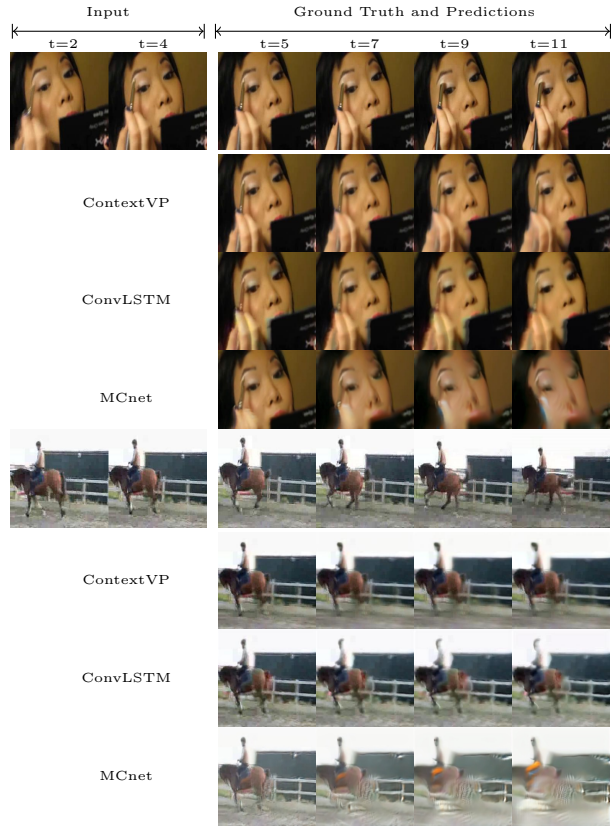

\section{Conclusion and Future Directions}
\label{sec:conclusion}
This paper identified the issue of missing context in current video prediction models, which contributes to uncertain predictions about the future and leads to generation of blurry frames.
To address this issue, we developed a novel prediction architecture that captures all of the relevant context efficiently at each layer.
It outperformed existing approaches for video prediction in a variety of scenarios, demonstrating the importance of fully context-aware models.

We did not incorporate other recent ideas for improve video prediction such as explicit background/motion flow modeling, or adversarial training.
Since these have been previously explored for models with incomplete context, a promising future direction is to evaluate their influence on fully context-aware models.
Our work suggests that full context coverage should be a required feature of any video prediction baseline to rule out multiple sources of uncertainty.

\pagestyle{headings}
\mainmatter

\title{Supplemental Material \\ContextVP: Fully Context-Aware \\ Video Prediction} 



\author{}
\institute{}

\maketitle

\section{Results}
This section shows more visual examples of next-step and multi-step prediction results. 
Our best model compared below is 4 layer ContextVPs with weighted blending and directional weight sharing (ContextVP4-WD-big) and the baseline is a networks of 20 ConvLSTM layers with skip connections (ConvLSTM20) (See the main paper for more details of the network architecture).

\textbf{KITTI/Caltech dataset:} \autoref{fig:result-caltech} shows the comparison of next-step prediction results of our best model, the baseline, and Deep Predictive Coding Network (PredNet; \cite{lotter2016deep}).
All models are trained for next-frame prediction given 10 input frames on the KITTI dataset, and tested on the CalTech Pedestrian dataset.

\noindent\textbf{UCF-101 dataset:} \autoref{fig:result-ucf} compares multi-step predictions of our best model with the baseline and MCnet+RES \cite{villegas2017decomposing}.
All models are trained for next-frame prediction given 4 input frames and tested for 8 frames.
Apprearance of objects such as the barbell, the toothbrush, and the mop are preserved much better by ContextVP. 





\noindent\begin{figure}
\centering
\tiny
\begin{tabular}{  c @{\hspace{0.7em}} c @{\hspace{0.2em}} c @{\hspace{0.2em}} c @{\hspace{0.2em}} c }
    Input &  & \multicolumn{3}{c}{Predictions (t=11)} \\
      & Ground &  & & \\
     t=2 \quad \quad \quad t=5 \quad \quad \quad  \quad t=8  & Truth & ContextVP & ConvLSTM & PredNet \\

    \includegraphics[width=5.1cm]{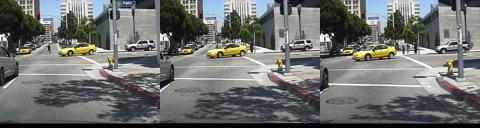} &
    \includegraphics[width=1.7cm]{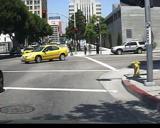} &
    \includegraphics[width=1.7cm]{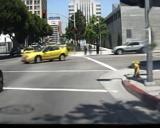} &
    \includegraphics[width=1.7cm]{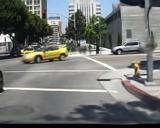} &
    \includegraphics[width=1.7cm]{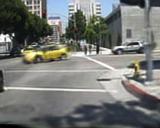} \\ [-1em]
    
    \includegraphics[width=5.1cm]{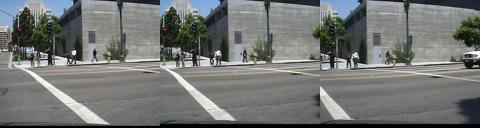} &
    \includegraphics[width=1.7cm]{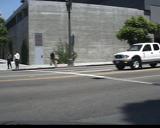} &
    \includegraphics[width=1.7cm]{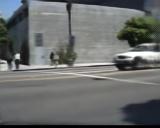} &
    \includegraphics[width=1.7cm]{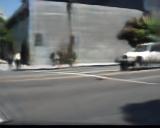} &
    \includegraphics[width=1.7cm]{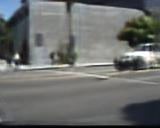} \\ [-1em]
    
    \includegraphics[width=5.1cm]{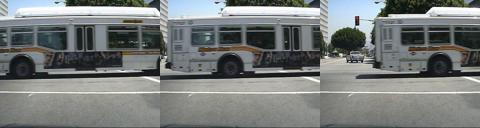} &
    \includegraphics[width=1.7cm]{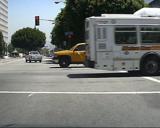} &
    \includegraphics[width=1.7cm]{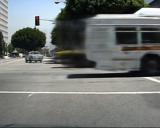} &
    \includegraphics[width=1.7cm]{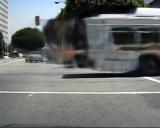} &
    \includegraphics[width=1.7cm]{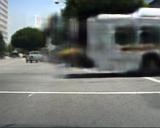} \\ [-1em]

    \includegraphics[width=5.1cm]{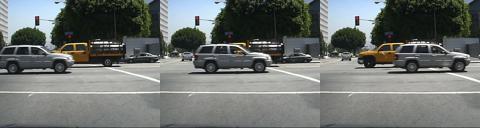} &
    \includegraphics[width=1.7cm]{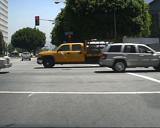} &
    \includegraphics[width=1.7cm]{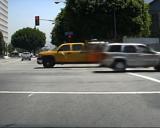} &
    \includegraphics[width=1.7cm]{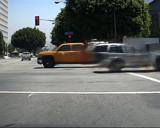} &
    \includegraphics[width=1.7cm]{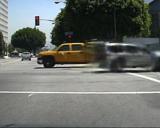} \\ [-1em]

    \includegraphics[width=5.1cm]{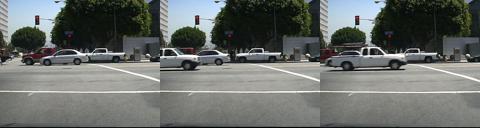} &
    \includegraphics[width=1.7cm]{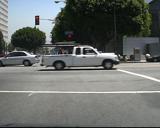} &
    \includegraphics[width=1.7cm]{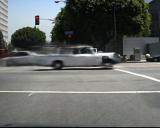} &
    \includegraphics[width=1.7cm]{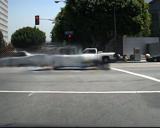} &
    \includegraphics[width=1.7cm]{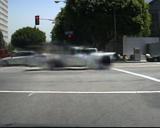} \\ [-1em]

    \includegraphics[width=5.1cm]{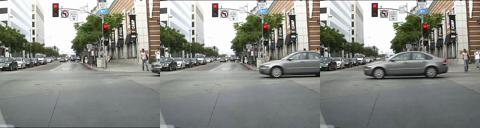} &
    \includegraphics[width=1.7cm]{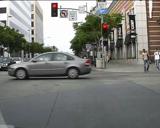} &
    \includegraphics[width=1.7cm]{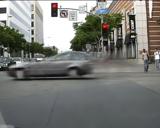} &
    \includegraphics[width=1.7cm]{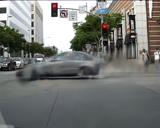} &
    \includegraphics[width=1.7cm]{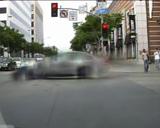} \\ [-1em]

    \includegraphics[width=5.1cm]{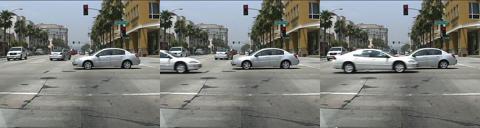} &
    \includegraphics[width=1.7cm]{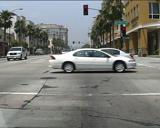} &
    \includegraphics[width=1.7cm]{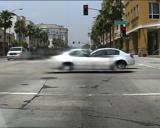} &
    \includegraphics[width=1.7cm]{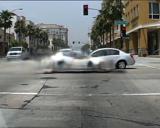} &
    \includegraphics[width=1.7cm]{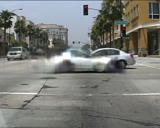} \\ [-1em]

    \includegraphics[width=5.1cm]{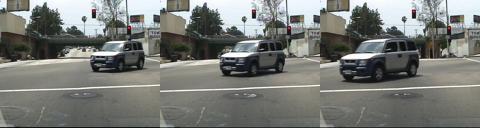} &
    \includegraphics[width=1.7cm]{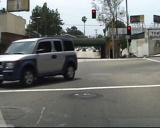} &
    \includegraphics[width=1.7cm]{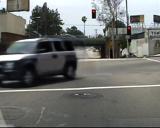} &
    \includegraphics[width=1.7cm]{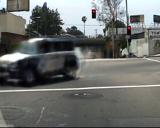} &
    \includegraphics[width=1.7cm]{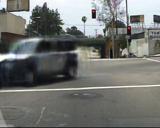} \\ 
  
\end{tabular}
\begin{tikzpicture}[overlay, remember picture, yshift=.25\baselineskip, shorten >=.5pt, shorten <=.5pt,align=left]
    \draw[|<->|] (-6.1,10) -- (-0.9,10);
    \draw[|<->|] (0.9,10) -- (6.2,10);
  \end{tikzpicture}
\caption{Qualitative comparisons from the test set among our best model (ContextVP4-WD-big), the baseline (ConvLSTM20), and the state-of-the-art model (PredNet). All models are trained for next-frame prediction given 10 input frames on the KITTI dataset, and tested on the CalTech Pedestrian dataset.}
\label{fig:result-caltech}
\end{figure}

\noindent\begin{figure}
\centering
\begin{tabular}{ c  @{\hspace{0.3em}} c @{\hspace{0.0em}} c @{\hspace{0.3em}}  @{\hspace{0.3em}} c @{\hspace{0.0em}} c  @{\hspace{0.0em}}}
    {\scriptsize time} &  {\scriptsize$T-1$} \hspace{1.1cm} {\scriptsize$T$}  & {\scriptsize$T+1$} & {\scriptsize$T-1$} \hspace{1.1cm} {\scriptsize$T$} & {\scriptsize$T+1$} \\ [-0.2cm]
    &
    \multirow{2}*{\imagetop{\includegraphics[width=3.5cm]{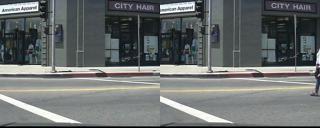}}} &
    \multirow{2}*{\imagetop{\includegraphics[width=1.7cm]{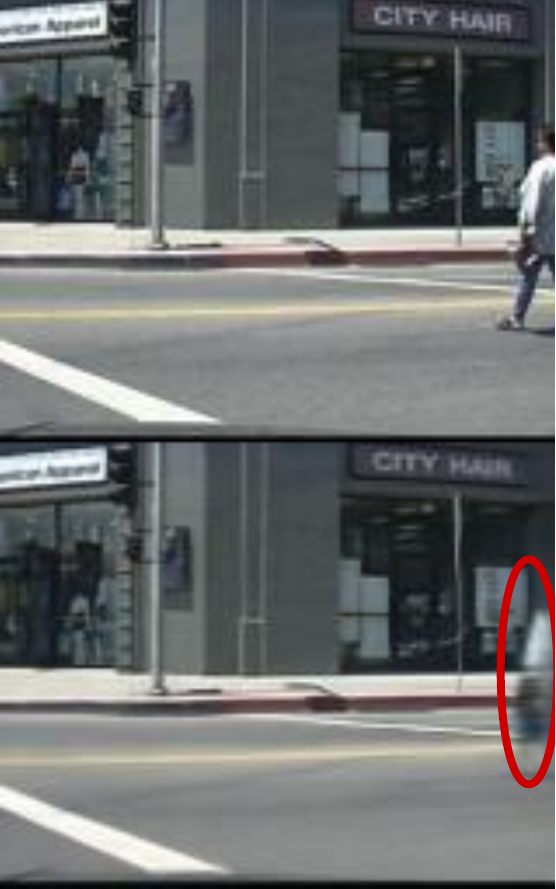}}} &
    \multirow{2}*{\imagetop{\includegraphics[width=3.5cm]{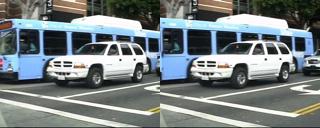}}} &
    \multirow{2}*{\imagetop{\includegraphics[width=1.7cm]{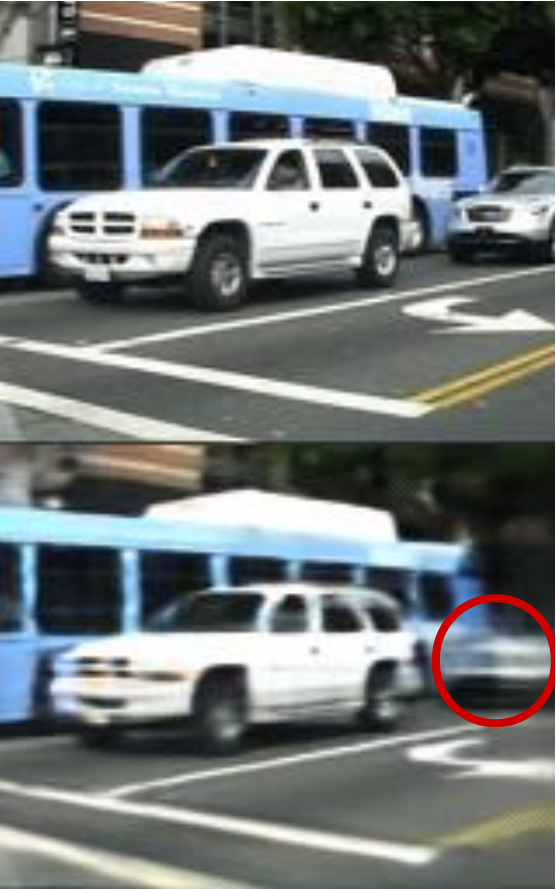}}}  \\ [0.5cm]
    {\scriptsize true} & & & & \\ [1.0cm]
    {\scriptsize predicted} & & & & \\ 

\end{tabular}
\vspace{0.8cm}
\caption{Examples of failed prediction for car-cam video sequences. The first rows are the true frames and the second rows are predictions. $T-1$ and $T$ are the last two input frames, and $T+1$ is the current output frame. As can be seen in the predicted frame at time $T+1$, some details of objects which did not appear in the last frames, are missed.}
\label{fig:failure}
\end{figure}

\noindent\begin{figure}
\centering
\tiny
\begin{tabular}{  c @{\hspace{0.7em}} c }
    Input & Ground Truth and Predictions\\
     & \\ 
    t=2 \quad \quad \quad t=4 &  t=5 \quad \quad \quad t=7 \quad \quad \quad t=9 \quad \quad \quad t=11 \\

    \includegraphics[width=2.8cm]{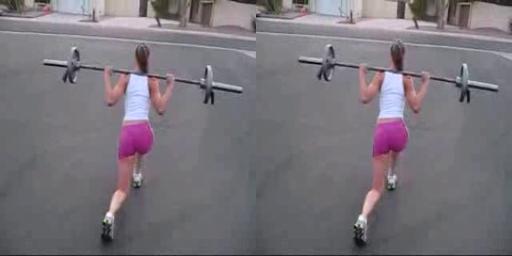} &
    \includegraphics[width=5.6cm]{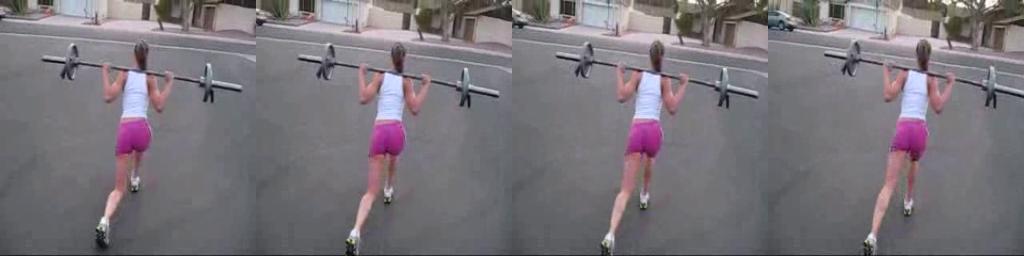} \\
    \quad \quad ContextVP & \includegraphics[align=c, width=5.6cm]{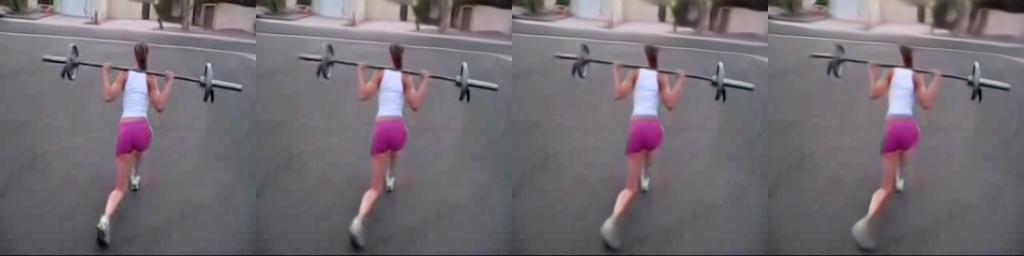} \\
    \quad \quad ConvLSTM & \includegraphics[align=c, width=5.6cm]{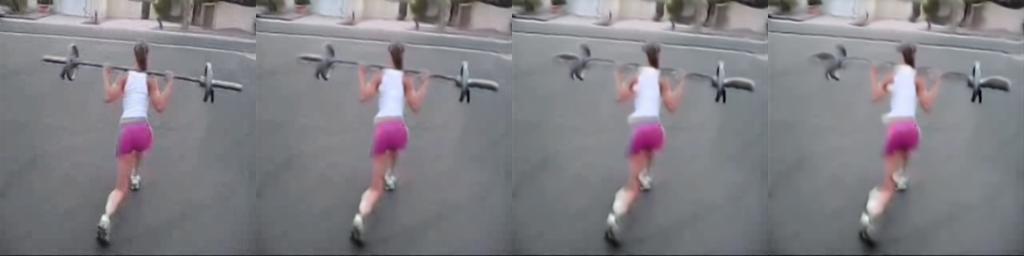} \\
    \quad \quad MCnet & \includegraphics[align=c, width=5.6cm]{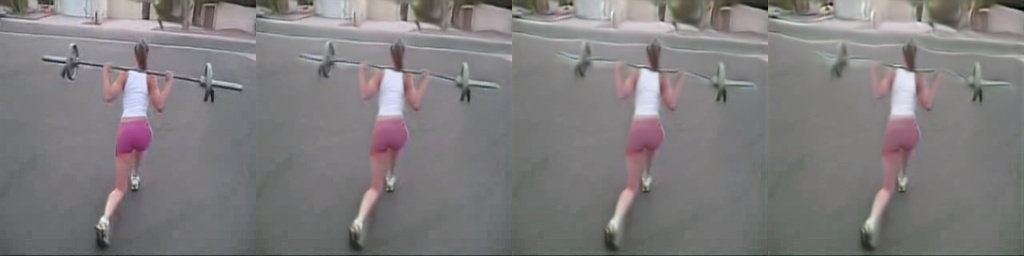} \\ [-1em]

    \includegraphics[width=2.8cm]{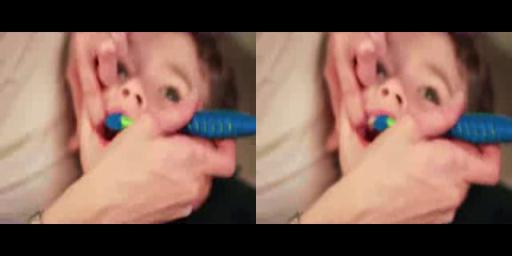} &
    \includegraphics[width=5.6cm]{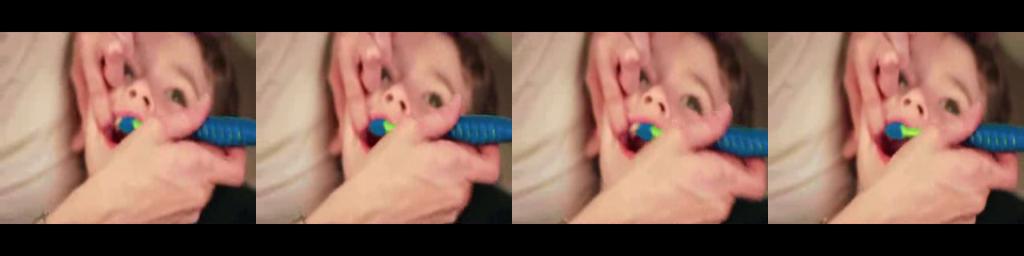} \\
    \quad \quad ContextVP & \includegraphics[align=c, width=5.6cm]{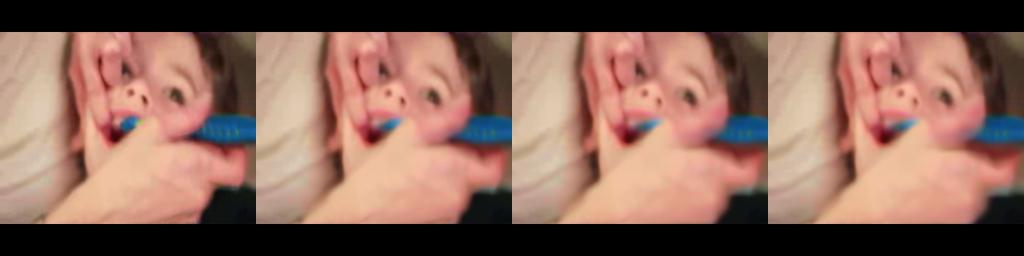} \\
    \quad \quad ConvLSTM & \includegraphics[align=c, width=5.6cm]{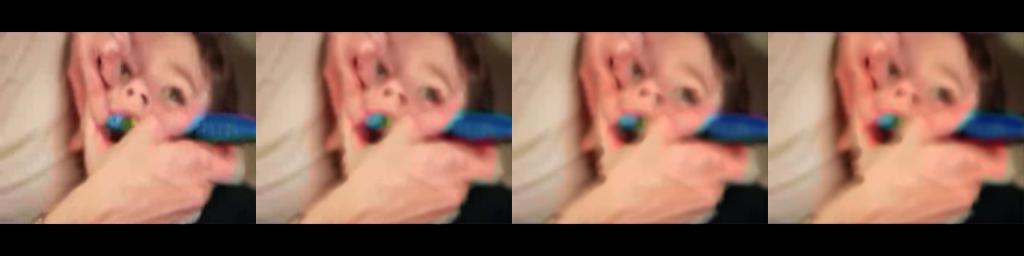} \\
    \quad \quad MCnet & \includegraphics[align=c, width=5.6cm]{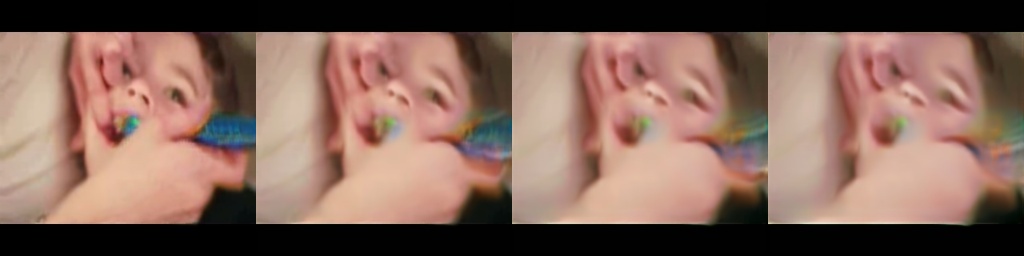} \\ [-1em]

    \includegraphics[width=2.8cm]{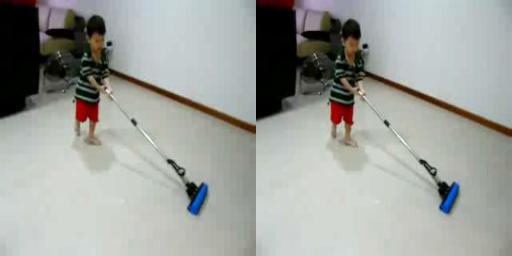} &
    \includegraphics[width=5.6cm]{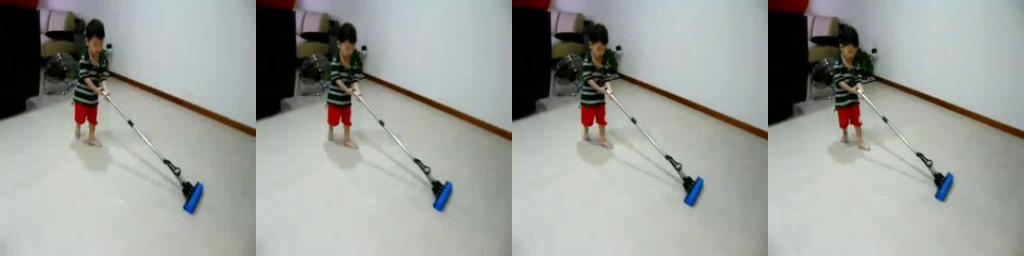} \\
    \quad \quad ContextVP & \includegraphics[align=c, width=5.6cm]{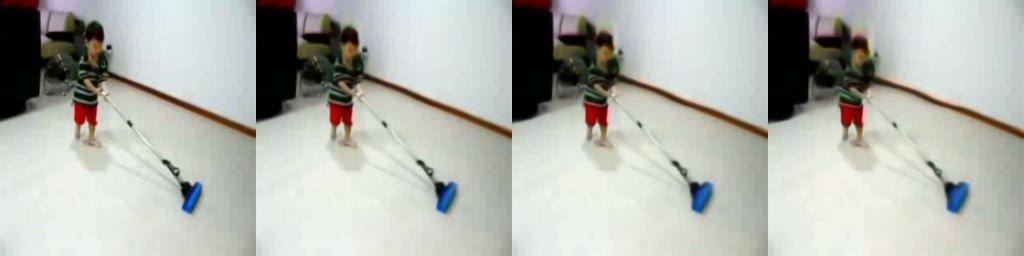} \\
    \quad \quad ConvLSTM & \includegraphics[align=c, width=5.6cm]{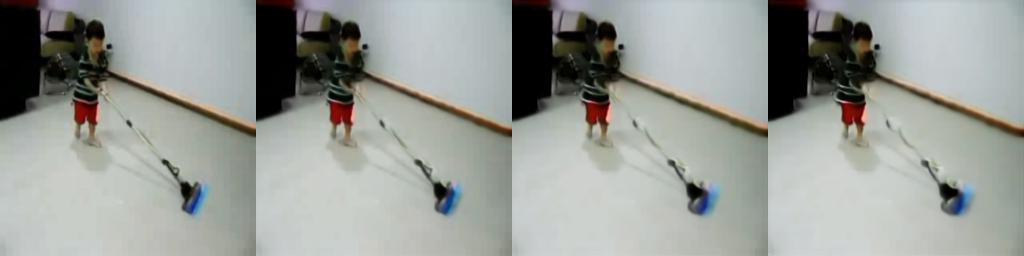} \\
    \quad \quad MCnet & \includegraphics[align=c, width=5.6cm]{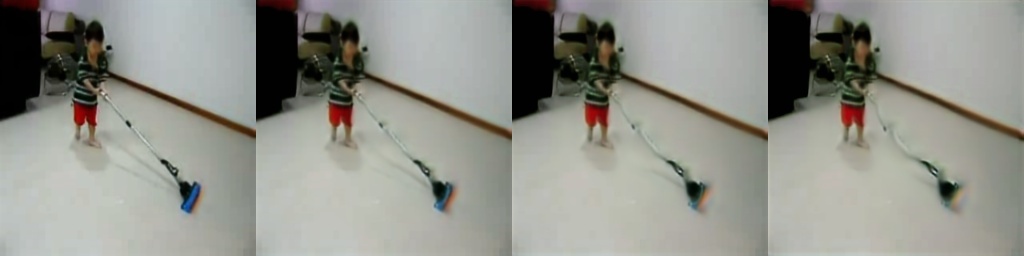} \\

\end{tabular}
\begin{tikzpicture}[overlay, remember picture, yshift=.25\baselineskip, shorten >=.5pt, shorten <=.5pt,align=left]
    \draw[|<->|] (-8.7,8.2) -- (-5.9,8.2);
    \draw[|<->|] (-5.7,8.2) -- (-0.1,8.2);
  \end{tikzpicture}
\caption{Qualitative comparisons from the UCF-101 test set among our best model (ContextVP4-WD-big), the baseline (ConvLSTM20), and the state-of-the-art model (MCNet). All models are trained for next-frame prediction given 4 input frames. They are then tested to recursively predict 8 future frames.}
\label{fig:result-ucf}
\end{figure}

\section{Failure Cases}
\autoref{fig:failure} shows two examples where our model fails to make clear predictions, in particular when objects are entering the frame. 
Although most of objects and background in the predicted frame at time $T+1$ look sharp, the objects marked with a red circle appear blurry. 
These particular objects hardly appear in the last (input) frames before prediction (see true frames at time $T-1$, $T$ in \autoref{fig:failure}), so the input does not have sufficient information to predict details of the object (see the predicted frames at time $T+1$ in \autoref{fig:failure}). 
%
%
%
\bibliographystyle{splncs04}
\bibliography{bib}

\end{document}